\documentclass{article}

\PassOptionsToPackage{numbers, compress}{natbib}
\usepackage[final]{neurips_2025}

\usepackage[utf8]{inputenc} % allow utf-8 input
\usepackage[T1]{fontenc}    % use 8-bit T1 fonts
\usepackage{hyperref}       % hyperlinks
\usepackage{url}            % simple URL typesetting
\usepackage{booktabs}       % professional-quality tables
\usepackage{amsfonts}       % blackboard math symbols
\usepackage{nicefrac}       % compact symbols for 1/2, etc.
\usepackage{microtype}      % microtypography
\usepackage{xcolor}         % colors
\usepackage{graphicx}
\usepackage{caption} 
\usepackage{subcaption}
\usepackage{tabularx}
\usepackage{pifont}

\usepackage{amsmath}
\usepackage{amssymb}
\usepackage{mathtools}
\usepackage{amsthm}
\usepackage{multirow}
\usepackage{colortbl}

\usepackage[textsize=tiny]{todonotes}
\usepackage{tcolorbox}

\definecolor{uparrowcolor}{RGB}{235, 255, 235}   % 浅绿色
\definecolor{downarrowcolor}{RGB}{255, 235, 235} % 浅红色

\newcommand{\uparrowcolor}[1]{\cellcolor{uparrowcolor}\textcolor{black}{$\uparrow$#1}}
\newcommand{\downarrowcolor}[1]{\cellcolor{downarrowcolor}\textcolor{black}{$\downarrow$#1}}

\title{Atomic Thinking of LLMs: Decoupling and Exploring Mathematical Reasoning Abilities}

\author{%
  Jiayi Kuang$^1$, Haojing Huang$^2$, Yinghui Li$^{2,\dagger}$\thanks{Yinghui Li is the project leader.}, Xinnian Liang$^3$, Zhikun Xu$^4$ \\
  \textbf{Yangning Li}$^2$, \textbf{Xiaoyu Tan}$^5$, \textbf{Chao Qu}$^6$, \textbf{Meishan Zhang}$^7$, \textbf{Ying Shen}$^{1,8}$\thanks{Correspond to Yinghui Li (liyinghuihhh@gmail.com), Ying Shen (sheny76@mail.sysu.edu.cn). Ying Shen is the corresponding author.}, \textbf{Philip S. Yu}$^9$ \\
  $^1$Sun Yat-sen University, $^2$Tsinghua University, $^3$ByteDance Inc.\\
  $^4$Arizona State University, $^5$ Tencent Youtu Lab, $^6$Fudan University\\
  $^7$Harbin Institute of Technology (Shenzhen)\\ $^8$Guangdong Provincial Key Laboratory of Fire Science and Intelligent Emergency Technology\\ $^9$University of Illinois Chicago
}

\begin{document}

\maketitle

\begin{abstract}
  Large Language Models (LLMs) have demonstrated outstanding performance in mathematical reasoning capabilities. However, we argue that current large-scale reasoning models primarily rely on scaling up training datasets with diverse mathematical problems and long thinking chains, which raises questions about whether LLMs genuinely acquire mathematical concepts and reasoning principles or merely remember the training data. In contrast, humans tend to break down complex problems into multiple fundamental atomic capabilities. Inspired by this, we propose a new paradigm for evaluating mathematical atomic capabilities. Our work categorizes atomic abilities into two dimensions: (1) \textbf{field-specific abilities across four major mathematical fields}, algebra, geometry, analysis, and topology, and (2) \textbf{logical abilities at different levels}, including conceptual understanding, forward multi-step reasoning with formal math language, and counterexample-driven backward reasoning. We propose corresponding training and evaluation datasets for each atomic capability unit, and conduct extensive experiments about how different atomic capabilities influence others, to explore the strategies to elicit the required specific atomic capability. Evaluation and experimental results on advanced models show many interesting discoveries and inspirations about the different performances of models on various atomic capabilities and the interactions between atomic capabilities. Our findings highlight the importance of decoupling mathematical intelligence into atomic components, providing new insights into model cognition and guiding the development of training strategies toward a more efficient, transferable, and cognitively grounded paradigm of ``\emph{atomic thinking}''. 
\end{abstract}

\section{Introduction}\label{sec:introduction}

In recent years, as Large Language Models (LLMs) have achieved remarkable performance in language understanding \cite{zhenglmsys,DBLP:conf/acl/LiZLLLSWLCZ22,wu2023symbol,DBLP:conf/nips/LiZLML0HY24,DBLP:conf/emnlp/DuW0D0LZVZSZGL024,DBLP:journals/tkde/LiHZZLLCZS23,DBLP:journals/csur/DongLGCLSY23}, visual perception \cite{DBLP:conf/icassp/WangLL0JK25,DBLP:journals/csur/KuangSXLXLLCLH25,hu2024bliva,DBLP:conf/acl/KuangL0L0J25,zhou2025humanvbenchexploringhumancentricvideo,yue2024less,jiao2025imgdiff,DBLP:conf/acl/LiXC0LMJLZZS24}, complex reasoning \cite{DBLP:journals/corr/abs-2507-09477,havrilla2024glore,tan2025struct,yuanadvancing,DBLP:journals/corr/abs-2403-12077,2025arXiv250805128Y} agentic intelligence \cite{DBLP:conf/acl/LuoKLS0D25,islam2024mapcoder,wang2024gta,DBLP:journals/corr/abs-2501-01945} and honesty \cite{tan2023self,chujie2024honestllm,hua2024trustagent,DBLP:conf/iclr/LiHKLGQTZSY25}, mathematical reasoning has emerged as a key focus for LLM cognitive abilities \cite{geminiteam2024gemini, yang2024qwen2,luo2023wizardmath, yumetamath}. Mathematics, as a fundamental reasoning task, offers both verifiable answers and a wide range of difficulty levels, drawing increasing research attention \cite{shao2024deepseekmath,ying2024internlm, abel}. Recent studies, particularly those involving reasoning models such as OpenAI's o1 and DeepSeek-R1, have demonstrated strong mathematical performance, achieving impressive results on many challenging benchmarks \cite{openai2023gpt4,DBLP:conf/iclr/LiLWJZZWZH0Y25,DBLP:conf/coling/XuLD0CJZLXH25}.

Current approaches enhance mathematical performance by scaling up training data, incorporating diverse mathematical problems, and complex reasoning paradigms \cite{yu2023metamath, li2024common, yang2024qwen2}. WizardMath \cite{luo2023wizardmath} synthesizes complex multi-field data via varied math instructions, while models like OpenAI o1 train extended chains of thought to perform higher-level reasoning. As the complexity of mathematical problems increases, some studies, such as Lean4 \cite{Lean4} and Lean-STaR \cite{lin2024lean}, leverage formal mathematical languages to mitigate ambiguity in natural language multi-step reasoning, as well as enhancing the reward model for problem-solving process \cite{DBLP:conf/acl/SunDCZ25,tan2025aurora,DBLP:conf/nips/DidolkarGKGVLRB24}. However, as model size and data scale approach saturation, these training paradigms are facing challenges:
\begin{itemize}
    \item \textit{Do models truly grasp mathematical concepts and inference patterns, or are they merely memory problems by chain of thought reasoning training?}
    \item \textit{Is there a more fundamental cognitive atom that can break through the current paradigm, and \textbf{what are the advantages of this atomic thinking compared with the chain of thought?}}
\end{itemize}

\begin{figure}[t]
    \centering
    \includegraphics[width=1\linewidth]{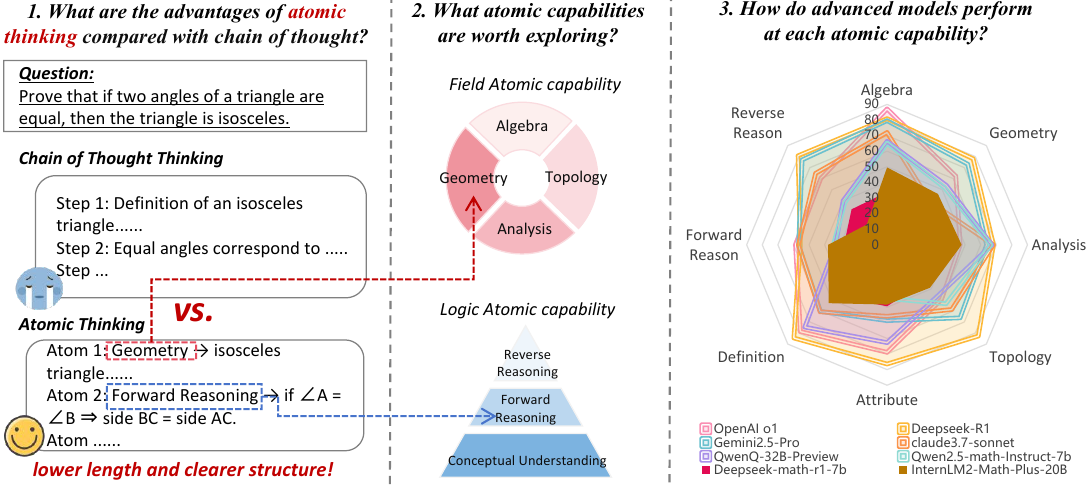}
    \caption{This figure illustrates an overview of our atomic thinking. It compares thought chains and atomic thinking, highlighting the efficiency of atomic thinking. Next, it shows the atomic capabilities we focus on. Finally, it provides the performance of advanced models in every atomic capability.}
    \label{fig:intro}
\end{figure}

Figure \ref{fig:intro} compares the differences between the two paradigms. Existing math reasoning strategies, such as chain-of-thought and tree-of-thought, tend to rely heavily on sequential context \cite{shao2024deepseekmath,mirzadeh2024gsm, yu2024reasonagain, luo2024chain}. This not only leads to inefficient use of computational resources but also introduces noise through excessive self-correction in long reasoning chains. In contrast, human reasoning typically decomposes complex problems into atomic problems, solving them incrementally and integrating only the essential information for subsequent steps, which is referred to as \textit{atomic thinking} \cite{teng2025atom}. This atomic thinking paradigm promotes flexible and structural reasoning, enabling more efficient problem-solving like data probe \cite{ling2025diversity,DBLP:conf/coling/HuangMLHZ0Z24,DBLP:conf/aaai/YuJLHWLCLLTZZXH24}. Thus, decoupling the mathematical atomic abilities of LLMs is not only essential for assessing their cognitive depth but also key to transitioning from the current ``question drilling'' paradigm to a \textit{``atomic thinking''} framework. So we wonder \textbf{\textit{what atomic capabilities are worth exploring?}}

Current mathematical benchmarks mostly assess models' accuracy in end-to-end problem solving \cite{amini2019mathqa,liu2024mathbench, lu2024mathvista}, offering little insight into a systematic assessment of atomic capabilities. To address this, we propose a novel framework for exploring mathematical atomic abilities, encompassing both field and logical reasoning capabilities, with an emphasis on ensuring minimal overlap between different atomic units while maintaining broad coverage of mathematical tasks. For field atomic abilities, we draw inspiration from modern mathematics and construct four foundational fields: \textbf{algebra, geometry, analysis, and topology}. For logical reasoning abilities, we reference cognitive psychology in math reasoning to define three core capabilities: (1) \textbf{Conceptual Understanding}, which grasps math definitions and axioms; (2) \textbf{Forward Reasoning with Formal Math Language}, which conducts rigorous, multi-step reasoning using symbolic systems; (3) \textbf{Counterexample-driven Backward Reasoning}, requiring constructing counter-examples and leveraging backward reasoning. For each atomic ability, we construct training and testing data, ensuring interpretability and isolating cross-ability. We conduct the evaluation experiments as shown in Figure \ref{fig:intro} to explore: \textbf{\textit{how do advanced models perform at each atomic capability?}}

Beyond decoupled atomic capability evaluation, we also conduct composite experiments across reasoning levels and varied atomic abilities to investigate how atomic abilities influence each other. Our findings on all the results reveal several key insights:

\begin{itemize}
    \item \textbf{Field-level performance}: LLMs perform better in algebra and analysis, while struggling in geometry and topology. Interestingly, models exhibit atypical behavior in topology, performing worse on easier tasks yet better on harder ones.

    \item \textbf{Logical reasoning abilities}: Larger LLMs exhibit stronger conceptual understanding, likely due to superior pretraining memory. However, even advanced commercial models struggle with constructing counterexamples, indicating a gap in backward reasoning skills.

    \item \textbf{Cross-field interaction}: Training on low-difficulty data can hinder high-level skill expression in some fields. Notably, activating algebraic abilities significantly improves performance in other fields, which is often more than direct training in the target field.

    \item \textbf{Cross-Logic interaction}: Conceptual understanding enhances other reasoning abilities and field atomic abilities. Surprisingly, training solely on definition completion tasks suffices to stimulate high-level ability, outperforming models trained on more complex data. This reveals the supporting value of conceptual understanding in mathematical training.
\end{itemize}

\section{Related work}
\paragraph{Mathematics-enhanced large language models}
Mathematical reasoning has become a key focus in exploring the upper bounds of LLMs’ cognitive capabilities~\cite{DBLP:journals/corr/abs-2405-12819,DBLP:journals/corr/abs-2404-04925}. Unlike general-purpose models such as GPT-4 \cite{openai2023gpt4} and Gemini \cite{geminiteam2024gemini}, math-enhanced LLMs emphasize field-specific strategies, including data augmentation, pretraining, fine-tuning \cite{wei2022chain}, and reinforcement learning on large-scale mathematical corpora \cite{abel}. \textbf{WizardMath} \cite{luo2023wizardmath} synthesizes diverse math data through instruction generation and leverages RLHF and process supervision. \textbf{NuminaMath} \cite{numina_math} adopts Tool-Integrated Reasoning (TIR) to generate math data, including questions with fine-grained solutions. \textbf{Qwen2.5-Math} \cite{yang2024qwen2} fine-tunes Qwen2.5 on proprietary high-quality math data. \textbf{InternLM2-Math} \cite{ying2024internlm} enhances logical rigor by integrating formal mathematical language, code interpreters, and theorem proving in Lean4. \textbf{Deepseek-Math-rl} \cite{shao2024deepseekmath} focuses on data engineering and efficient RL training.

\paragraph{Mathematical benchmarking}
Due to their objective correctness and structured difficulty~\cite{DBLP:journals/corr/abs-2307-09007,DBLP:journals/corr/abs-2402-11420}, math tasks are ideal for evaluating LLMs \cite{amini2019mathqa, yang2019learning, zhengminif2f, liu2024mathbench, lu2024mathvista}. Benchmarks like \textbf{MATH} \cite{hendrycksmath2021} and \textbf{GSM8K} \cite{cobbe2021gsm8k} test high school and elementary-level reasoning and have become standard. To meet the demands of advanced models, more challenging datasets such as \textbf{OlympiadBench} \cite{he2024olympiadbench} target competition-level exams. Formal theorem proving benchmarks like \cite{tsoukalasputnambench} \textbf{Putnam Bench}, \textbf{CoqGym} \cite{yang2019learning}, and \textbf{MiniF2F} \cite{zhengminif2f} further assess logical reasoning with tools like Coq and Lean. However, most benchmarks assess end-to-end question-solving performance without decomposing tasks into \textbf{atomic reasoning abilities}. Our work aims to decouple and analyze these atomic abilities and their interactions to support finer-grained reasoning evaluation and lightweight but effective training.

\section{Atomic capability decoupling and interaction}

Inspired by the atomization of human cognition, we propose decoupling the mathematical capabilities of LLMs into atomic abilities. We categorize atomic capabilities into two major types, ensuring that each capability is disentangled from the others while jointly covering a wide range of mathematical tasks. We construct corresponding training and test sets to evaluate the model performance of different atomic abilities. Beyond evaluating individual capabilities, we further investigate their interactions. Specifically, we explore how stimulating one atomic capability may affect others, offering insights into how such interactions can be leveraged to enhance targeted atomic abilities and promote compositional problem-solving strategies.

\subsection{Atomic capability design}

\paragraph{Field atomic capabilities}
We refer to the core field division of modern mathematics. It is worth noting that if we divide the capabilities too finely (e.g., dividing them into more than ten types), the atomic units will be too loose and the interaction between atomic abilities will be too complex, which may not be a particularly significant correlation. Therefore, we finally divide the field's atomic abilities into \textbf{Algebra}, \textbf{Geometry}, \textbf{Analysis}, and \textbf{Topology}. To analyze the influence of difficulty levels in a more fine-grained aspect, we further divide each field into low difficulty (level 1) and high difficulty (level 2). Similarly, we divide it into two difficulty levels to avoid complex interactions.

\paragraph{Logical reasoning atomic capabilities}
For logic reasoning atomic ability, we refer to the human cognitive mode, deconstruct the general process of complex reasoning, and identify core abilities:
\begin{enumerate}
    \item \textbf{Conceptual understanding} is a fundamental, which includes \emph{definition identification} and \emph{attribute description}. The definition identification task requires the LLM to complete the name of the corresponding definition in a statement. The property description task requires completing the detailed description in a mathematical definition, including the premises, the conditions, the key words, and parameters.
    \item \textbf{Forward reasoning with formal language}: With reference to human cognitive processes, we designate forward multi-step reasoning as a higher-level logical atomic ability. Since natural language reasoning is often confronted with vague proofs and uncritical assumptions, we emphasize the forward reasoning using formal mathematical language.
    \item \textbf{Backward reasoning with counterexamples}: In addition to step-by-step forward multi-step reasoning, backward reasoning with counterexamples is also a very important ability in mathematical reasoning. By skillfully constructing appropriate counterexamples, proofs can be effectively accomplished that are difficult to perform directly with forward reasoning. 

\end{enumerate}
It is worth noting that we do not consider computational abilities. This is because we are mainly concerned with the logical reasoning ability of mathematical reasoning. Computational capability is difficult to decouple from the reasoning process. In addition, when computations are needed, LLM invoking relevant mature computational tools would be more efficient and accurate.

\begin{figure}[t]
    \centering
    \includegraphics[width=0.88\linewidth]{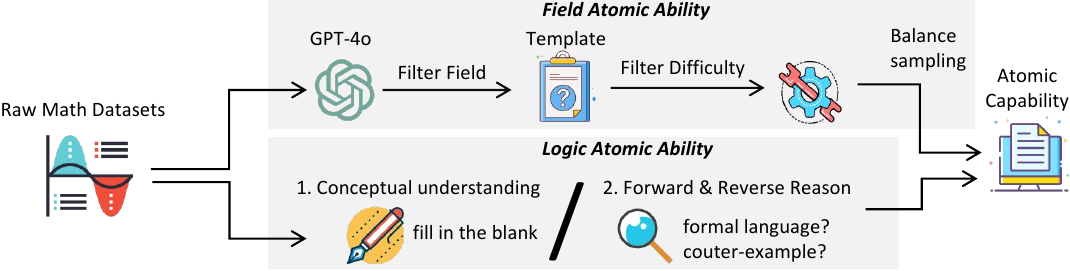}
    \caption{This figure illustrates our data construction procedure.}
    \label{fig:data_cons}
\end{figure}
\vspace{-3mm}

\subsection{Data construction for capability evaluation}
To decouple each atomic capability, we construct the training and testing data, which contributes to evaluation and further exploration of the interaction. The overview of our data construction is shown in Figure \ref{fig:data_cons}. More data statistics and examples can be found in Appendix \ref{app:data_statis}.

For field capabilities, we collect data from current benchmarks such as MATH \cite{hendrycksmath2021}, GSM8K \cite{cobbe2021gsm8k}, Gaokao-Bench \cite{gaokao-bench}, OlympiadBench \cite{he2024olympiadbench}, AIME, MMLU \cite{wang2024mmlu}, and DeepMath \cite{he2025deepmath}. Using a combination of template matching and LLM-assisted annotation, we reclassify problems into four fields. When there are original field labels in the raw data, LLM takes more account of the primitive labels to better align with human thinking. Difficulty annotation is done by template matching. We take into account the data sources of the original questions (e.g., primary and secondary school questions or Olympiad questions), the original difficulty labeling information, and we emphasize aligning the difficulty classifications of different fields. After that, we randomly sample the data to ensure that the number of each field is relatively balanced. Finally, we randomly divide the training set and test set with a ratio of 3:1. For \textbf{conceptual understanding}, we extract math definitions and axioms from NaturalProofs \cite{naturalproof} and generate fill-in-the-blank questions. For \textbf{forward reasoning}, we collect questions and proofs with formal language such as LeanWorkbook, and filter the questions with definite answers. For \textbf{backward reasoning}, we use a counterexample-driven reasoning statement from CounterMath \cite{li2025one}. Given data scarcity, we maintain a 1:1 train-test split for logic atomic capability to ensure evaluation robustness.

\subsection{Atomic capability interaction}
In addition to decoupling individual atomic capabilities, we also try to explore the correlation and interaction between atomic capabilities. One of the most basic strategies is trying to stimulate a certain atomic capability and observe whether it has an impact on other atomic capabilities. This will provide insight for subsequent research on how to specifically stimulate the required atomic capabilities. Therefore, we explore the following atomic capability interactions:

\begin{itemize}
    \item \textbf{Cross-difficulty Interaction:} We investigate how training on Level 1 or Level 2 tasks within a field affects performance at the other level.

    \item \textbf{Cross-field Interaction:} We train on tasks from one field and evaluate transfer effects to others, particularly among related or complementary fields.

    \item \textbf{Logical Capability Interactions:} We examine whether conceptual understanding can support high-level ability, improving forward or backward reasoning, and whether forward and backward reasoning mutually reinforce each other. Additionally, we assess whether high-level reasoning enhances foundational understanding.

    \item \textbf{Reasoning-to-field Interaction:} Given the abstract nature of some fields (e.g., topology and analysis), we test whether improving logical reasoning capabilities can enhance performance in field-specific tasks, especially those requiring conceptual abstraction.
\end{itemize}

\section{Experimental settings}\label{sec:experiment}
\subsection{Decoupled atomic capability evaluation} \label{metric}

\paragraph{Baselines} 
We evaluate a diverse set of LLMs with a focus on mathematical reasoning. Open-source models include \textbf{Deepseek-Math-7B-RL} \cite{shao2024deepseekmath}, \textbf{Eurus-2-7B-PRIME} \cite{cui2024process}, \textbf{Qwen2.5-Math-7B-Instruct} \cite{yang2024qwen2}, \textbf{NuminaMath-7B-TIR} \cite{numina_math}, \textbf{InternLM2-Plus-7B/20B} \cite{ying2024internlm}, \textbf{Abel-7B/13B} \cite{abel}, \textbf{WizardMath-7B} \cite{luo2023wizardmath}, \textbf{Mathstral-7B}\footnote{\url{https://mistral.ai/news/mathstral/}}, \textbf{MetaMath-Mistral-7B} \cite{yu2023metamath}, \textbf{Xwin-Math-7B/13B} \cite{li2024common}, \textbf{QwQ-32B}\footnote{\url{https://qwenlm.github.io/blog/qwq-32b-preview/}}. For \textit{proprietary models}, we use \textbf{GPT-4o}, \textbf{OpenAI o1},\footnote{\url{https://cdn.openai.com/o1-system-card-20241205.pdf}}, \textbf{Deepseek-R1}\footnote{\url{https://api-docs.deepseek.com/news/news250120}}, \textbf{Claude3.7-sonnet}\footnote{\url{https://www.anthropic.com/claude/sonnet}}, and \textbf{Gemini2.5-pro}\footnote{\url{https://deepmind.google/technologies/gemini/pro/}}. This selection spans varied training paradigms, data sources, architectures, and origins (academic vs industrial). Open-source evaluations are conducted on 4$\times$ L20 48GB GPUs, while proprietary models are accessed via official APIs.

\paragraph{Prompts and metrics} 
We adopt default \textit{Chain-of-Thought} (CoT) prompting, instructing models to enclose answers in \texttt{\textbackslash boxed\{\}} for extraction. Accuracy is computed via exact match with reference answers. For \textbf{Conceptual Understanding}, models complete missing definitions. \textbf{Forward Reasoning} requires multi-step derivations using formal mathematical language. Accuracy is used for both. For \textbf{Backward Reasoning}, we refer to previous work CounterMATH \cite{li2025one}: (1) F-1 score on the statement judgement, and (2) Example, Strict, Loose metric to evaluate whether the model constructs valid counterexamples. Detailed description of our prompts and metrics can be found in Appendix \ref{app:prompt}.

\subsection{Training for atomic capability interaction}

To examine interactions among atomic abilities, we fine-tune \texttt{Qwen2.5-Math-Instruct-7B} using supervised LoRA training~\cite{hu2022lora} on 4$\times$L20 48GB GPUs, with a learning rate of 1.0e-5. Training about different interactions adheres to a unified hardware setup and consistent hyperparameters. After training, evaluation follows Section~\ref{metric} to ensure comparability across settings.

\begin{table}[t]
    \centering
    \caption{Model performance on field atomic capabilities. We \textbf{bold} the optimal and \underline{underline} the suboptimal of models. The low and high difficulty levels correspond to level 1 and level 2, respectively.}
    \vspace{2.5mm}
    \label{tab:field_performance}
    \resizebox{\linewidth}{!}{
    \begin{tabular}{l l c c c c c c c c}
        \toprule[1.3pt]
        \multicolumn{2}{c}{Field} & \multicolumn{2}{c}{Algebra} & \multicolumn{2}{c}{Geometry} & \multicolumn{2}{c}{Analysis} & \multicolumn{2}{c}{Topology} \\
        \cmidrule(lr){3-4} \cmidrule(lr){5-6} \cmidrule(lr){7-8} \cmidrule(lr){9-10}
        \multicolumn{2}{c}{Difficulty Level} & Low & High & Low & High & Low & High & Low & High \\
        \midrule[1.3pt]
        \multicolumn{10}{c}{\textit{\textbf{Open-source models}}} \\
        \midrule[1.3pt]
        \multirow{12}{*}{Model$\leq$7B} & InternLM2-math-plus-7b & 49.2 & 35.9 & 33.0 & 31.4 & 41.9 & 41.5 & 27.2 & 37.0 \\
        & Deepseek-math-rl-7b & 52.0 & 33.8 & 35.5 & \underline{39.3} & 44.6 & 37.6 & 22.6 & 33.7 \\
        & Eurus-2-7B-PRIME & 50.7 & 33.9 & 36.5 & 32.3 & 45.4 & 32.0 & \underline{36.5} & 23.3 \\
        & NuminaMath-7B-TIR & 52.4 & 28.2 & 39.9 & 32.1 & \underline{46.8} & 33.6 & 26.4 & 25.1 \\
        & MetaMath-Mistral-7B & \underline{59.6} & \underline{41.4} & \underline{42.4} & 37.7 & 45.2 & \underline{37.9} & 20.6 & 25.8 \\
        & Mathstral-7B-v1.0 & 42.3& 33.3& 36.4& 28.1& 32.9& 29.8& 24.1& \underline{37.1} \\
        & Abel-7B-002 & 46.7 & 32.9 & 36.1 & 35.2 & 32.9 & 26.3 & 25.6 & 31.5 \\
        & Xwin-Math-7B & 52.6 & 37.3 & 38.3 & 35.1 & 45.3 & 32.9 & 19.7 & 22.5 \\
        & Qwen2.5-math-Instruct-7b & \textbf{80.5} & \textbf{65.2} & \textbf{57.3} & \textbf{51.9} & \textbf{67.7} & \textbf{66.5} & \textbf{52.1} & \textbf{53.4} \\
        \midrule
        \multirow{4}{*}{Model > 7B} & Abel-13B & 63.4 & 47.3 & 42.7 & 41.8 & 40.4 & 35.6 & 30.2 & 24.6 \\
        & Xwin-Math-13B & 78.3 & 51.6 & \underline{49.5} & \underline{46.7} & \underline{55.7} & \underline{54.8} & \underline{46.7} & 38.3 \\
        & IntwenLM2-Math-Plus-20B & \underline{67.8} & \underline{49.6} & 49.3 & 46.1 & 54.3 & 47.8 & 45.4 & \underline{38.9} \\
        & QwQ-32B-Preview & \textbf{85.1} & \textbf{66.3} & \textbf{59.6} & \textbf{53.8} & \textbf{72.3} & \textbf{67.2} & \textbf{54.9} & \textbf{46.1} \\
        \midrule[1.3pt]
        \multicolumn{10}{c}{\textit{\textbf{Commercial models}}} \\
        \midrule[1.3pt]
        & OpenAI o1 & \textbf{93.6}& \textbf{87.1}& 66.3& 62.3& 59.3& 46.9& 52.9 & 49.4\\
        & GPT-4o & 69.3&50.5&48.5&36.3&55.1&53.2&53.0 &49.4\\
        & Deepseek-r1 & \underline{83.6}& \underline{80.9}& \textbf{76.8}& \textbf{78.0} & \underline{70.3}& \textbf{68.2}&\textbf{77.7} &\textbf{82.7} \\
        & Claude3.7-sonnet & 83.5&72.0&58.0&39.0&\textbf{80.3}&\underline{68.1}&56.0&58.6 \\
        & Gemini2.5-pro & 80.5& 79.6& \underline{70.2}& \underline{73.1}& 67.5& 63.8&\underline{68.1} & \underline{66.2}\\
        \bottomrule[1.3pt]
    \end{tabular}}
\end{table}

\section{Analysis and discussion}

\subsection{Experimental analysis of field atomic capabilities}

We evaluate several advanced models by assessing their performance across decoupled atomic abilities in distinct mathematical fields. The detailed results are presented in Table \ref{tab:field_performance}. We observe that:

\paragraph{Larger models exhibit stronger atomic capabilities}
Model performance varies significantly across scales. In general, larger models perform better, benefiting from greater capacity and more extensive training data. Among 7B-scale models, \textbf{Qwen2.5-math-Instruct} achieves notably superior results across all evaluated fields, even outperforming some larger models such as \textbf{InternLM2-Math-Plus-20B}. Analysis of its outputs shows that it generates longer reasoning chains, facilitating deeper logical inference and enhancing its problem-solving capabilities.

\paragraph{Algebra and analysis perform better}
Models tend to exhibit stronger mathematical atomic abilities in Algebra and Analysis, while performance in Geometry and especially Topology remains weaker. Since we analyze the training data of the open-sourced models we have evaluated, the results reveal that Geometry and Topology are significantly underrepresented in the training data. Moreover, although geometric problems are textually presented, they often require spatial or visual reasoning—an area where LLMs typically struggle. Consequently, current models demonstrate limited atomic capabilities in these fields. Improving performance in underrepresented areas, particularly Topology and Geometry, is a pressing research challenge. For example, the low performance on topology-related tasks may reflect a lack of understanding of abstract mathematical structures. A promising direction is to integrate core mathematical concepts during training to stimulate relevant atomic skills. We explore such cross-ability interactions in Section \ref{exp:interactions}.

\paragraph{Internal difficulty-level analysis}
We further examine model performance across different difficulty levels within each field. In Algebra, models often experience sharp drops in accuracy on high-difficulty problems, indicating a gap in advanced atomic capabilities. In contrast, performance degradation in Geometry and Analysis is less severe. This suggests the need for training paradigms that better balance basic and advanced skill acquisition to ensure robust generalization. An interesting anomaly arises in Topology, where models sometimes perform better on harder problems than on easier ones. We hypothesize that this is due to mismatches between the models’ training distributions and our evaluation data: some high-difficulty problems may incidentally align with abstract patterns the models have implicitly learned. This counterintuitive result shows that what humans find hard may not be hard for LLMs, which encourages deeper exploration into the field atomic ability decomposition.

Further results in Appendix \ref{app:difficulty} show that training data difficulty significantly affects model performance. Notably, excessively low-difficulty training data may degrade accuracy across difficulty levels. Thus, balancing training data difficulty is essential for fostering generalizable atomic capabilities.

\begin{table}[t]
    \centering
    \caption{Model performance on logic atomic capabilities, where Attr. and Def. are short names of the Attribute description and definition task. We \textbf{bold} the optimal and \underline{underline} the suboptimal results.}
    \vspace{2.5mm}
    \label{tab:logic_performance}
    \resizebox{\linewidth}{!}{
    \begin{tabular}{l l c c c c c c c}
        \toprule
        \multicolumn{2}{c}{} & \multicolumn{2}{c}{Concept} & \multicolumn{1}{c}{Forward Rea.} & \multicolumn{4}{c}{Backward Rea.} \\
        \cmidrule(lr){3-4} \cmidrule(lr){5-5} \cmidrule(lr){6-9}
        \multicolumn{2}{c}{} & Attr. (Acc.) & Def. (Acc.) & Acc. & F-1 & Example(\%) & Strict(\%) & Loose(\%) \\
        \midrule[1.3pt]
        \multicolumn{9}{c}{\textit{\textbf{Open-source models}}} \\
        \midrule[1.3pt]
        \multirow{9}{*}{Model$\leq$7B} & InternLM2-math-plus-7b & \textbf{43.2} & 46.2 & 23.7 & 33.9 & 36.6 & 9.0 & 9.5 \\
        & Deepseek-math-rl-7b & \underline{39.4} & \underline{46.5} & 27.6 & 32.2 & 65.9 & 18.9 & 20.6 \\
        & Eurus-2-7B-PRIME & 23.1 & 27.9 & \textbf{41.4} & \underline{37.5} & 64.8 & \underline{28.5} & \underline{32.0} \\
        & NuminaMath-7B-TIR & 22.8 & 27.3& 30.2 & 30.4 & 54.1 & 13.0 & 13.7 \\
        & MetaMath-Mistral-7B & 19.6 & 25.6 & 28.6 & 31.0 & 26.5 & 0.4 & 0.7 \\
        & Mathstral-7B-v1.0 & 21.7 & 29.8 & 32.9 & 28.2 & 38.9 & 7.5 & 7.9 \\
        & Abel-7B-002 & 20.9 & 31.0 & 33.7 & 34.4 & \underline{66.1} & 16.0 & 17.9 \\
        & Xwin-Math-7B & 18.8 & 26.3 & 26.4 & 28.1 & 31.3 & 1.2 & 1.7 \\
        & Qwen2.5-math-Instruct-7b & 34.4 & \textbf{50.3} & \underline{34.4} & \textbf{38.3} & \textbf{74.2} & \textbf{30.2} & \textbf{33.2} \\
        \midrule
        \multirow{4}{*}{Model > 7B} 
        & Abel-13B &  31.2 & 48.0 & 37.2  & 22.4 & 24.4 & 0.8 & 0.8 \\
        & Xwin-Math-13B & 29.8 & 45.9 & 33.1 & \underline{30.2} & \underline{31.3} & 1.2 & 1.7 \\
        & InternLM2-Math-Plus-20B & \underline{38.3} & \underline{52.9} & \underline{37.8} & 18.4 & 28.8 & \underline{8.4} & \underline{9.5} \\
        & QwQ-32B &\textbf{62.7} & \textbf{74.6} & \textbf{42.6} & \textbf{39.9} & \textbf{70.0} & \textbf{38.6} & \textbf{43.8} \\
        \midrule[1.3pt]
        \multicolumn{9}{c}{\textbf{\textit{Commercial models}}} \\
        \midrule[1.3pt]
        & OpenAI o1 & \underline{68.9} & \underline{78.6} & \textbf{58.7} & 60.1 & 55.8 & 39.8 & 40.9 \\
        & GPT-4o & 38.1 & 48.3 & 37.7 & 59.0 & 44.7 & 19.7 & 21.3 \\
        & Deepseek-r1 & \textbf{76.4} & \textbf{84.5} & 55.6 & \textbf{80.7} & \underline{86.8} & \underline{54.2} & \underline{65.3} \\
        & Claude3.7-sonnet & 45.8 & 60.7 & \underline{56.8} & 64.8 & 78.0 & 45.0 & 52.5 \\
        & Gemini2.5-pro & 48.6 & 60.4 & 56.7 & \underline{77.0} & \textbf{90.8} & \textbf{65.1} & \textbf{75.7} \\
        \bottomrule[1.3pt]
    \end{tabular}}
\end{table}

\subsection{Experimental analysis of mathematical logical reasoning atomic capabilities}

We evaluate several state-of-the-art mathematical models across multiple dimensions of logical reasoning. Detailed results are shown in Table~\ref{tab:logic_performance}. From these results, there are some insights that:

\paragraph{Models recognize definitions but struggle with deeper conceptual understanding}
All models perform better at recognizing definitions than completing missing properties, indicating a surface-level grasp of mathematical rigor. This suggests that while models can identify known concepts, they often lack precise internal representations. Larger commercial models significantly outperform smaller open-source ones; for example, \texttt{deepseek-r1} scores 84.5 in definition recognition and 76.4 in property completion. This disparity reflects the importance of pretraining, where larger models benefit from superior long-range memory and MoE (mixture-of-experts) mechanisms that mitigate knowledge forgetting. Current math models emphasize problem-solving over conceptual understanding, contributing to the gap between basic concept recognition and advanced reasoning or proof tasks. Section~\ref{case_study} provides further case studies.

\paragraph{Structured reasoning with formal language remains a challenge for smaller models}
Since large-scale commercial models show better performance with the best accuracy of 58.7, the models with smaller parameters struggle with both understanding the questions with formal math language and applying formal mathematical language to reason, despite performing reasonably well in natural-language-based reasoning. This suggests that intensive ``problem-drilling'' may promote pattern memorization over structured formal reasoning. Some models attempt to bridge this gap by generating code-like representations to aid multi-step deduction. Notably, reasoning-oriented models such as \texttt{o1} and \texttt{DeepSeek-R1}, equipped with stronger long-range inference and self-reflection, achieve outstanding results in this category.

\paragraph{Limited counterexample abilities reveal the limits of problem-solving training}
For the ability to judge the truth value of mathematical statements, open-source models average around 30 F-1 points, and even advanced models like \texttt{QWQ-32B} reach only 39.9, while \texttt{Deepseek-R1}, optimized for mathematical reasoning, scores 80.7. In generating counterexamples, the \texttt{Qwen} series performs particularly well, sometimes surpassing commercial models like \texttt{o1}. Conversely, models like \texttt{MetaMath} succeed in only 26.5\% of such tasks. Although the best model Gemini2.5-pro demonstrates superior performance across various metrics when it constructs counter-examples, the performance of the other models on example consistency is still relatively low, with almost none exceeding 50\%. This reflects the limitations of training paradigms overly focused on direct problem-solving, which hinders higher-level abstraction and conceptual reasoning.

\subsection{Experimental analysis of interactions between atomic abilities} \label{exp:interactions}

\subsubsection{Influence between atomic abilities across different fields}

\begin{table}[t]
\centering
\caption{Performance comparison after stimulating various \textbf{\textit{field}} atomic capabilities. We color the \colorbox{uparrowcolor}{positive\uparrowcolor{}} / \colorbox{downarrowcolor}{negative\downarrowcolor{}} influence as \colorbox{uparrowcolor}{green} / \colorbox{downarrowcolor}{red}.}
\vspace{2.5mm}
\resizebox{\linewidth}{!}{
\begin{tabular}{l c c c c c c c c}
\toprule
Field & \multicolumn{2}{c}{Algebra} & \multicolumn{2}{c}{Analysis} & \multicolumn{2}{c}{Geometry} & \multicolumn{2}{c}{Topology} \\
\cmidrule(lr){2-3} \cmidrule(lr){4-5} \cmidrule(lr){6-7} \cmidrule(lr){8-9}
Difficulty Level & Low & High & Low & High & Low & High & Low & High \\
\midrule
Qwen-base & 80.5 & 65.2 & 67.7 & 66.5 & 52.1 & 53.4 & 52.1 & 53.4 \\
\midrule
Qwen-train-Algebra & 80.2 (\downarrowcolor{0.3}) & 69.7 (\uparrowcolor{4.5}) & 75.8 (\uparrowcolor{8.1}) & 71.5 (\uparrowcolor{5.0}) & 65.7 (\uparrowcolor{13.6}) & 57.5 (\uparrowcolor{4.1}) & 56.0 (\uparrowcolor{3.9}) & 62.3 (\uparrowcolor{8.9}) \\
Qwen-train-Analysis & 81.3 (\uparrowcolor{0.8}) & 66.4 (\uparrowcolor{1.2}) & 71.8 (\uparrowcolor{4.1}) & 64.9 (\downarrowcolor{1.6}) & 58.2 (\uparrowcolor{6.1}) & 55.8 (\uparrowcolor{2.4}) & 46.1 (\downarrowcolor{6.0}) & 54.1 (\uparrowcolor{0.7}) \\
Qwen-train-Geometry & 79.6 (\downarrowcolor{0.9}) & 68.1 (\uparrowcolor{2.9}) & 69.8 (\uparrowcolor{2.1}) & 59.5 (\downarrowcolor{7.0}) & 57.3 (\uparrowcolor{5.2}) & 56.0 (\uparrowcolor{2.6}) & 52.8 (\uparrowcolor{0.7}) & 60.9 (\uparrowcolor{7.5}) \\
Qwen-train-Topology & 79.7 (\downarrowcolor{0.8}) & 64.4 (\downarrowcolor{0.8}) & 71.5 (\uparrowcolor{3.8}) & 59.2 (\downarrowcolor{7.3}) & 61.7 (\uparrowcolor{9.6}) & 55.6 (\uparrowcolor{2.2}) & 53.7 (\uparrowcolor{1.6}) & 59.1 (\uparrowcolor{5.7}) \\
\bottomrule
\end{tabular}}
\label{tab:field_interaction}
\end{table}

\begin{table}[t]
\centering
\caption{Performance comparison after stimulating various \textbf{\textit{field}} atomic capabilities that are trained on InternLM2-math-plus-7B.}
\vspace{2.5mm}
\resizebox{\linewidth}{!}{
\begin{tabular}{l c c c c c c c c}
\toprule
Field & \multicolumn{2}{c}{Algebra} & \multicolumn{2}{c}{Analysis} & \multicolumn{2}{c}{Geometry} & \multicolumn{2}{c}{Topology} \\
\cmidrule(lr){2-3} \cmidrule(lr){4-5} \cmidrule(lr){6-7} \cmidrule(lr){8-9}
Difficulty Level & Low & High & Low & High & Low & High & Low & High \\
\midrule
InternLM2-math-plus-7B & 49.2 & 35.9 & 33.0 & 31.4 & 41.9 & 41.5 & 27.2 & 37.0 \\
\midrule
InternLM2-train-Algebra & 48.1 (\downarrowcolor{1.1}) & 37.8 (\uparrowcolor{1.9}) & 35.8 (\uparrowcolor{2.8}) & 33.6 (\uparrowcolor{2.2}) & 43.1 (\uparrowcolor{1.2}) & 42.1 (\uparrowcolor{0.6}) & 27.8 (\uparrowcolor{0.6}) & 37.9 (\uparrowcolor{0.9}) \\
InternLM2-train-Geometry & 48.6 (\downarrowcolor{0.6}) & 36.3 (\uparrowcolor{0.4}) & 32.2 (\downarrowcolor{0.8}) & 30.5 (\downarrowcolor{0.8}) & 44.3 (\uparrowcolor{2.4}) & 44.0 (\uparrowcolor{2.5}) & 28.7 (\uparrowcolor{1.5}) & 38.8 (\uparrowcolor{1.8}) \\
\bottomrule
\end{tabular}}
\label{tab:field_interaction_Int}
\end{table}

We investigate the influence \textbf{ between fields} among field atomic abilities, as summarized in Table~\ref{tab:field_interaction}. In particular, atomic abilities in \textit{ algebra} consistently exhibit \textbf{positive effects} across all other fields, with particularly strong gains observed in \textit{Analysis} and \textit{Geometry}. In some cases, Algebra training even yields greater improvements than in-field training. For instance, when evaluating \textit{Geometry} atomic abilities, in-field training improved performance by 6.1 and 2.4 points at Levels 1 and 2, respectively, whereas activating Algebraic abilities led to larger gains of 13.6 and 4.2 points. This suggests a \textbf{complementary relationship} among atomic abilities, likely because algebraic problem-solving emphasizes fundamental reasoning skills that underpin more abstract fields, and our case study in Section \ref{case_study} demonstrates that. These findings highlight the potential of leveraging \textbf{diverse, field-specific atomic abilities} to enhance target capabilities more effectively. However, we also observe \textbf{negative transfer} effects. Strengthening atomic abilities in \textit{Topology} led to performance declines in \textit{Algebra} and \textit{Analysis}, with the largest drops reaching 0.8 and 7.3 points on Level 2. This may be due to substantial \textbf{data distribution divergence}. We have provided a visualization result in Appendix \ref{app:heatmap}. To further validate our conclusions, we have now included InternLM2-math-plus-7B as an additional baseline for training and evaluation, aiming to further strengthen our claims. The results, shown in Table \ref{tab:field_interaction_Int}, are consistent with our observations on Qwen2.5-math-instruct and provide a strong complement to our previous findings. These findings underscore the need to consider that, when trying to stimulate one, interactions between fields can significantly impact performance in unintended ways.

\begin{table}[t]
\centering
\caption{Performance comparison after stimulating various \textbf{\textit{logic}} atom capabilities.}
\vspace{2.5mm}
\resizebox{0.77\linewidth}{!}{
\begin{tabular}{l c c c c}
\toprule
Atom Capability & \multicolumn{2}{c}{Concept} & Formal language & Counter example \\
\cmidrule(lr){2-3} \cmidrule(lr){4-4} \cmidrule(lr){5-5}
 & Attribute & Definition & Acc. & F-1 \\
\midrule
Qwen-base & 34.4 & 50.3 & 34.4 & 30.2 \\
\midrule
Qwen-train-Concept & 34.8 (\uparrowcolor{0.4}) & 53.7 (\uparrowcolor{3.4}) & 53.5 (\uparrowcolor{19.1}) & 40.1 (\uparrowcolor{9.9}) \\
Qwen-train-Backward & 30.3 (\downarrowcolor{4.1}) & 46.3 (\downarrowcolor{4.0}) & 50.3 (\uparrowcolor{15.9}) & 41.1 (\uparrowcolor{10.9}) \\
Qwen-train-Forward & 25.1 (\downarrowcolor{9.3}) & 44.4 (\downarrowcolor{5.9}) & 53.7 (\uparrowcolor{19.3}) & 40.2 (\uparrowcolor{10.0}) \\
\bottomrule
\end{tabular}}
\label{tab:logic_capability_interaction}
\end{table}
\vspace{-3mm}

\subsubsection{Interactions among logical reasoning atomic abilities}

We also explore the interaction among \textbf{logical reasoning atomic abilities} as shown in Table \ref{tab:logic_capability_interaction}. Our findings indicate that \textit{conceptual understanding} plays a fundamental role in supporting two higher-level reasoning abilities. Training solely on fill-in-the-blank tasks that activate conceptual comprehension atomic ability results in substantial improvements of 19.1 and 9.9 in forward and backward reasoning. This aligns with our earlier analysis: current mathematical models, especially those with \textbf{limited parameter capacity}, exhibit significant deficiencies in conceptual understanding, which constrain the model’s capacity to develop more advanced reasoning skills. Furthermore, forward reasoning and counterexample construction appear to \textbf{mutually reinforce} each other, indicating the potential for bidirectional enhancement between forward and backward reasoning atomic abilities. However, enhancing high-level atomic abilities alone can lead to a decline in conceptual understanding, revealing that \textbf{overemphasis on solving complex problems} may lead to \textbf{forgetting of basic mathematical concepts}.

\begin{table}[t]
\centering
\caption{Performance comparison after stimulating conceptual understanding of atomic capabilities.}
\vspace{2.5mm}
\resizebox{\linewidth}{!}{
\begin{tabular}{l c c c c c c c c}
\toprule
 Field & \multicolumn{2}{c}{Algebra} & \multicolumn{2}{c}{Analysis} & \multicolumn{2}{c}{Geometry} & \multicolumn{2}{c}{Topology} \\
\cmidrule(lr){2-3} \cmidrule(lr){4-5} \cmidrule(lr){6-7} \cmidrule(lr){8-9}
Difficulty Level & Low & High & Low & High & Low & High & Low & High\\
\midrule
Qwen-base & 80.5 & 65.2 & 67.7 & 66.5 & 52.1 & 53.4 & 52.1 & 53.4 \\
\midrule
Qwen-train-Concept & 81.3 (\uparrowcolor{0.8}) & 66.1 (\uparrowcolor{0.9}) & 69.7 (\uparrowcolor{2.0}) & 72.3 (\uparrowcolor{5.8}) & 62.3 (\uparrowcolor{10.2}) & 59.8 (\uparrowcolor{6.4}) & 55.6 (\uparrowcolor{3.5}) & 57.2 (\uparrowcolor{3.8}) \\
\bottomrule
\end{tabular}}
\label{tab:field_concept_all}
\end{table}

\subsubsection{Influence of conceptual understanding on field-specific atomic abilities}

We also study the impact of logical atomic abilities on different fields. Particularly, we investigated how \textbf{conceptual understanding} supports field atomic abilities. The results in Table \ref{tab:field_concept_all} show \textbf{significant performance improvements}, particularly in fields that rely on abstract reasoning, such as \textit{Analysis and Geometry}. It is worth noting that activating conceptual understanding ability produces better outcomes than direct training on field-specific tasks. These results further confirm the \textbf{crucial role of conceptual understanding} as a foundational atomic ability, which highlights the need for future research to focus more on fostering \textbf{deep mathematical reasoning} and \textbf{concept-based learning}, rather than relying on \textbf{question repetition or difficulty escalation}.

\begin{figure}
    \centering
    \includegraphics[width=0.95\linewidth]{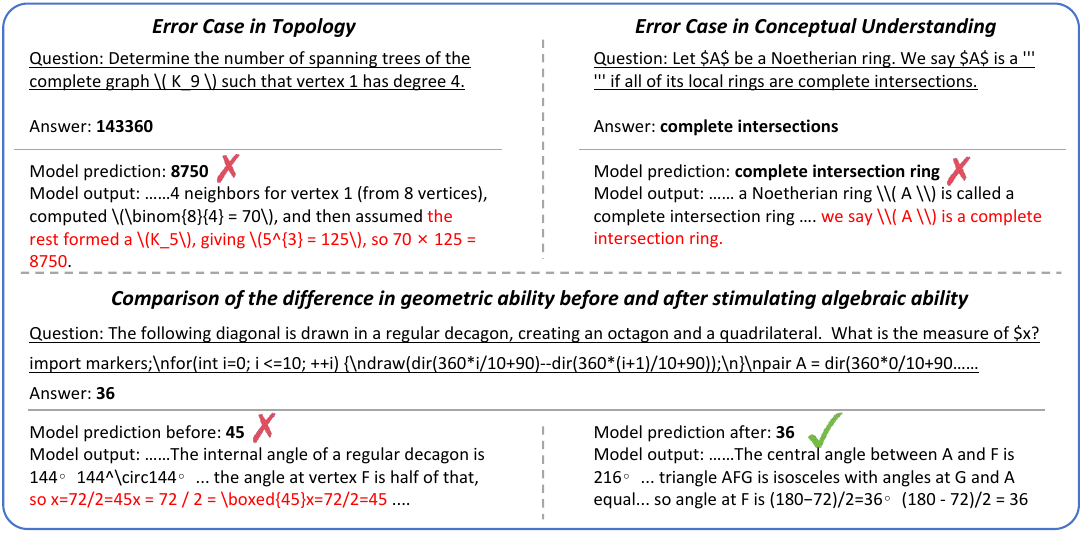}
    \caption{Case study about the error cases in Topology and Conceptual Atomic ability and the comparison of stimulating one atomic capability on another atomic capability.}
    \label{fig:case}
\end{figure}

\subsection{Case study}\label{case_study}
To better understand the deficiencies of the model in some atomic capabilities and the impact of stimulating one atomic capability on another atomic capability, we conduct a case study. Specifically, we compare model predictions before and after stimulating algebra ability on a geometry task. As shown in Figure \ref{fig:case}, before training, the model produces an incorrect result due to a flawed geometric assumption. After training, the model correctly applies knowledge of central angles and isosceles triangles, demonstrating the correlation from algebraic reasoning to geometric analysis.

\section{Conclusion, limitation, and future directions} \label{sec:conclusion}
In our work, we focus on decoupling the mathematical reasoning ability of LLM into atomic abilities and exploring the interactions between atomic abilities. We design field atomic capabilities and logical reasoning atomic capabilities with data. The evaluation results and analysis of decoupled atomic abilities on advanced models highlight the limited performance of some abilities. We further explore the interactions, for which we observe whether other atomic capabilities are affected when one atomic capability is stimulated. This process provides very interesting inspirations, including the facilitating role of algebraic abilities and the supporting of conceptual comprehension ability, both on logical and field atomic abilities. However, we have not explored more advanced strategies to stimulate a specific atomic capability, such as curriculum learning or reinforcement learning. Our findings will encourage exploring how to better stimulate the required atomic abilities and utilize multiple atomic abilities to solve complex mathematical tasks.

\section*{Acknowledgement}

This research is supported by Key-Area Research and Development Program of Guangdong Province, Granted No. 2024B1111060004. This research is also supported by Basic Research Fund of Shenzhen City (JCYJ20240813112009013).

\bibliographystyle{ieeetr}
\bibliography{reference}

\section*{NeurIPS Paper Checklist}

\begin{enumerate}

\item {\bf Claims}
    \item[] Question: Do the main claims made in the abstract and introduction accurately reflect the paper's contributions and scope?
    \item[] Answer: \answerYes{} % Replace by \answerYes{}, \answerNo{}, or \answerNA{}.
    \item[] Justification: Our main claims well reflect the motivation and contribution of this paper and provides inspiration for subsequent research.

\item {\bf Limitations}
    \item[] Question: Does the paper discuss the limitations of the work performed by the authors?
    \item[] Answer: \answerYes{} % Replace by \answerYes{}, \answerNo{}, or \answerNA{}.
    \item[] Justification: We discuss the limitations and future directions in Section \ref{sec:conclusion}.

\item {\bf Theory assumptions and proofs}
    \item[] Question: For each theoretical result, does the paper provide the full set of assumptions and a complete (and correct) proof?
    \item[] Answer: \answerNA{} % Replace by \answerYes{}, \answerNo{}, or \answerNA{}.
    \item[] Justification: Our work mainly explores the atomic capabilities of LLM mathematics reasoning from an experimental perspective as an empirical study, and does not conduct theoretical research.

    \item {\bf Experimental result reproducibility}
    \item[] Question: Does the paper fully disclose all the information needed to reproduce the main experimental results of the paper to the extent that it affects the main claims and/or conclusions of the paper (regardless of whether the code and data are provided or not)?
    \item[] Answer: \answerYes{} % Replace by \answerYes{}, \answerNo{}, or \answerNA{}.
    \item[] Justification: We provide the models, hardware information, prompt design, etc. used for training and evaluation in Section \ref{sec:experiment} and the Appendix \ref{app:prompt}, and provide our data and code in the supplementary materials for review. We will further disclose the relevant data and code to facilitate subsequent researchers to reproduce our results.

\item {\bf Open access to data and code}
    \item[] Question: Does the paper provide open access to the data and code, with sufficient instructions to faithfully reproduce the main experimental results, as described in supplemental material?
    \item[] Answer: \answerYes{} % Replace by \answerYes{}, \answerNo{}, or \answerNA{}.
    \item[] Justification: We provide our data and code in the supplementary materials for review. We will further disclose the relevant data and code in GitHub and Huggingface to facilitate subsequent researchers to reproduce our results.

\item {\bf Experimental setting/details}
    \item[] Question: Does the paper specify all the training and test details (e.g., data splits, hyperparameters, how they were chosen, type of optimizer, etc.) necessary to understand the results?
    \item[] Answer: \answerYes{} % Replace by \answerYes{}, \answerNo{}, or \answerNA{}.
    \item[] Justification: We provide the models, hardware information, prompt design, etc. used for training and evaluation in Section \ref{sec:experiment} and the Appendix \ref{app:prompt}.

\item {\bf Experiment statistical significance}
    \item[] Question: Does the paper report error bars suitably and correctly defined or other appropriate information about the statistical significance of the experiments?
    \item[] Answer: \answerNo{} % Replace by \answerYes{}, \answerNo{}, or \answerNA{}.
    \item[] Justification: Due to the limitation of experimental resources (especially for expensive commercial models) and considering that these studies cannot support the main claims, we did not study the error bars, confidence intervals and other statistical data in our experiments. In addition, we have provided sufficient experimental details, data, and codes to ensure the reproducibility and credibility of the results.

\item {\bf Experiments compute resources}
    \item[] Question: For each experiment, does the paper provide sufficient information on the computer resources (type of compute workers, memory, time of execution) needed to reproduce the experiments?
    \item[] Answer: \answerYes{} % Replace by \answerYes{}, \answerNo{}, or \answerNA{}.
    \item[] Justification: Details of the compute resources used to conduct the experiments are provided in Section \ref{sec:experiment}.
    
\item {\bf Code of ethics}
    \item[] Question: Does the research conducted in the paper conform, in every respect, with the NeurIPS Code of Ethics \url{https://neurips.cc/public/EthicsGuidelines}?
    \item[] Answer: \answerYes{} % Replace by \answerYes{}, \answerNo{}, or \answerNA{}.
    \item[] Justification: We promise that we strictly follow the code of ethics of NeurIPS 2025.

\item {\bf Broader impacts}
    \item[] Question: Does the paper discuss both potential positive societal impacts and negative societal impacts of the work performed?
    \item[] Answer: \answerYes{} % Replace by \answerYes{}, \answerNo{}, or \answerNA{}.
    \item[] Justification: We describe our impact in the Section \ref{sec:introduction}, \ref{sec:experiment}, and \ref{sec:conclusion}, and declare that our research has no conflicts of interest and will not cause negative impact.
    
\item {\bf Safeguards}
    \item[] Question: Does the paper describe safeguards that have been put in place for responsible release of data or models that have a high risk for misuse (e.g., pretrained language models, image generators, or scraped datasets)?
    \item[] Answer: \answerNA{} % Replace by \answerYes{}, \answerNo{}, or \answerNA{}.
    \item[] Justification: We have not released models and data with high risk of abuse. All the models and data used in our research have been strictly risk-controlled and open-source.

\item {\bf Licenses for existing assets}
    \item[] Question: Are the creators or original owners of assets (e.g., code, data, models), used in the paper, properly credited and are the license and terms of use explicitly mentioned and properly respected?
    \item[] Answer: \answerYes{} % Replace by \answerYes{}, \answerNo{}, or \answerNA{}.
    \item[] Justification: The models and data we use are all official open source, or call the official API. All use complies with their license.

\item {\bf New assets}
    \item[] Question: Are new assets introduced in the paper well documented and is the documentation provided alongside the assets?
    \item[] Answer: \answerYes{} % Replace by \answerYes{}, \answerNo{}, or \answerNA{}.
    \item[] Justification: All the data and codes we processed are organized and documented, and submitted together with the supplementary materials for review. We will further open source these data and codes to allow more researchers to participate in the exploration.

\item {\bf Crowdsourcing and research with human subjects}
    \item[] Question: For crowdsourcing experiments and research with human subjects, does the paper include the full text of instructions given to participants and screenshots, if applicable, as well as details about compensation (if any)? 
    \item[] Answer: \answerNA{} % Replace by \answerYes{}, \answerNo{}, or \answerNA{}.
    \item[] Justification: Our paper does not involve crowdsourcing nor research with human subjects.

\item {\bf Institutional review board (IRB) approvals or equivalent for research with human subjects}
    \item[] Question: Does the paper describe potential risks incurred by study participants, whether such risks were disclosed to the subjects, and whether Institutional Review Board (IRB) approvals (or an equivalent approval/review based on the requirements of your country or institution) were obtained?
    \item[] Answer: \answerNA{} % Replace by \answerYes{}, \answerNo{}, or \answerNA{}.
    \item[] Justification: Our paper does not involve crowdsourcing nor research with human subjects.

\item {\bf Declaration of LLM usage}
    \item[] Question: Does the paper describe the usage of LLMs if it is an important, original, or non-standard component of the core methods in this research? Note that if the LLM is used only for writing, editing, or formatting purposes and does not impact the core methodology, scientific rigorousness, or originality of the research, declaration is not required.
    %this research? 
    \item[] Answer: \answerNA{} % Replace by \answerYes{}, \answerNo{}, or \answerNA{}.
    \item[] Justification: The core method development in our research does not involve LLMs as any important, original, or non-standard components.

\end{enumerate}

\newpage

\appendix
\section{Additional details about data}\label{app:data_statis}

After our data construction process, we collected the data corresponding to each decoupled atomic capability and divided it into training set and test set. The specific statistical results are as following Table \ref{tab:data_stats_field} and Table \ref{tab:data_statis_logic_cap}:

\begin{table}[h]
\centering
\caption{Data statistics of \textbf{FIELD} atomic capabilities.}
\vspace{2.5mm}
\label{tab:data_stats_field}
\resizebox{\linewidth}{!}{
\begin{tabular}{l c c c c c c c c}
\toprule
\multirow{2}{*}{Field Cap.} & \multicolumn{2}{c}{Algebra} & \multicolumn{2}{c}{Geometry} & \multicolumn{2}{c}{Analysis} & \multicolumn{2}{c}{Topology} \\
\cmidrule(lr){2-3} \cmidrule(lr){4-5} \cmidrule(lr){6-7} \cmidrule(l){8-9}
& Level 1 & Level 2 & Level 1 & Level 2 & Level 1 & Level 2 &  Level 1 & Level 2 \\
\midrule
Train/Test & 3813/1277 & 4517/1505 & 3351/1117 & 3391/1331 & 3276/1092 & 4077/1358 & 3336/1112 & 3176/1058 \\
\bottomrule
\end{tabular}}
\end{table}

\begin{table}[h]
\centering
\caption{Data statistics of \textbf{LOGICAL} atomic capabilities.}
\vspace{2.5mm}
\resizebox{\linewidth}{!}{
\begin{tabular}{ccccc}
\toprule
\multirow{2}{*}{Logic Cap.} & \multicolumn{2}{c}{Conceptual Understanding} & Forward Reasoning & Backward Reasoning \\
\cmidrule(lr){2-3} \cmidrule(lr){4-4} \cmidrule(lr){5-5}
& Attribute Description & Definition & Formal Language & Counter-example \\
\midrule
Train/Test & 1225/1217 & 1683/1661 & 1061/1032 & 1225/1217 \\
\bottomrule
\end{tabular}}
\label{tab:data_statis_logic_cap}
\end{table}

\begin{figure}[h]
    \centering
    \includegraphics[width=1\linewidth]{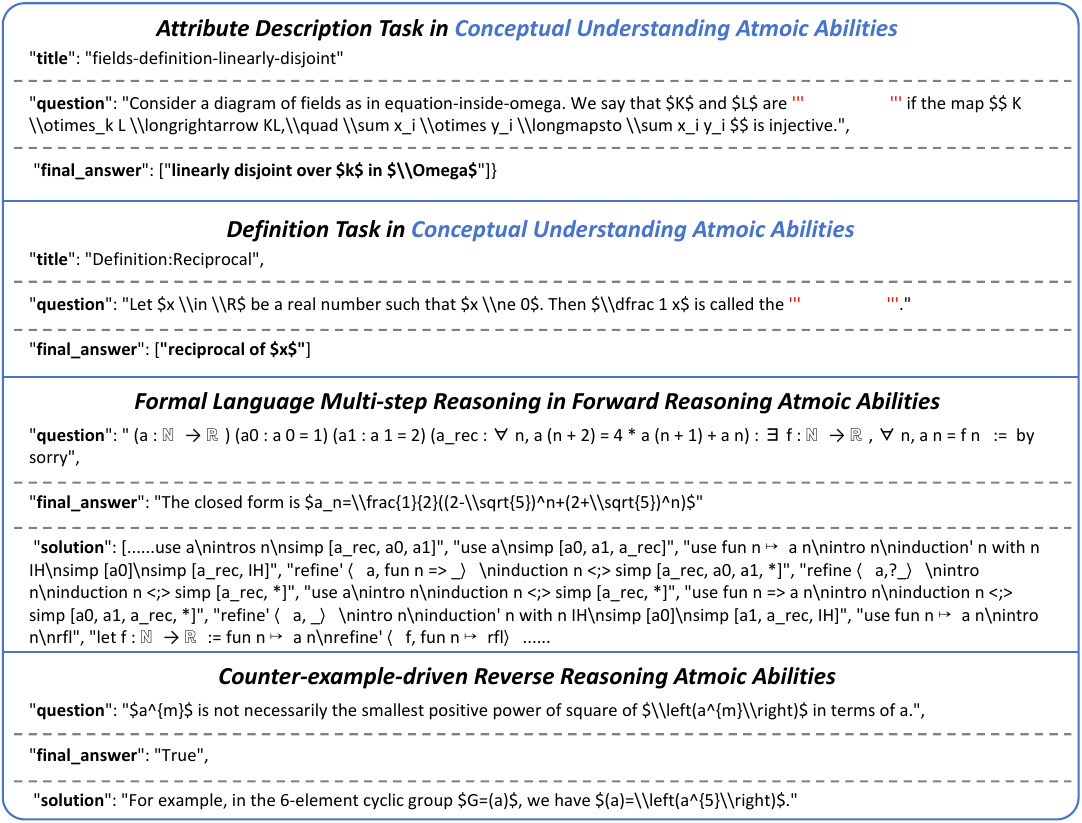}
    \caption{The figure presents data examples in logical atomic capability, including property descriptions, definition recognition, formal mathematical language-driven forward reasoning, and counterexample-driven backward reasoning.}
    \label{fig:data_example}
\end{figure}

In addition, we provide data examples of our logical atomic capabilities as shown in Figure \ref{fig:data_example}, to help readers better understand the different logical reasoning atomic tasks.

\section{Additional details about the experimental prompts and metrics}\label{app:prompt}
\paragraph{Prompt used for experiments}
The used prompts are summarized as follows. For field atomic capability, we use the same default chain of thought prompt:

\begin{tcolorbox}[title=\textit{Prompt for Field Atomic Capability},top=1mm,bottom=1mm]
\footnotesize
Please think step by step to solve the following question, and put your final answer within $\backslash\backslash$boxed\{\}.\\
\{question\}\\
\end{tcolorbox}

For conceptual understanding atomic capability, wo prompt the LLMs to fill in the blank:

\begin{tcolorbox}[title=\textit{Prompt for conceptual understanding Capability},top=1mm,bottom=1mm]
\footnotesize
Please think step by step to fill in the blank in '''  ''' of following statement, and put your final answer within $\backslash\backslash$boxed\{\}.\\
\{question\}\\
\end{tcolorbox}

For forward multi-step reasoning, we prompt the model to use formal math language to solve the question: 

\begin{tcolorbox}[title=\textit{Prompt for forward reasoning capability},top=1mm,bottom=1mm]
\footnotesize
Please think step by step to solve the following question by formal math language, and put your final answer within $\backslash\backslash$boxed\{\}.\\
\{question\}\\
\end{tcolorbox}

For counter-example-driven backward reasoning, we prompt the model to judge the statement true or false, which is the same setting of counter-math:

\begin{tcolorbox}[title=\textit{Counter-example backward reasoning prompt},top=1mm,bottom=1mm]
\footnotesize
\{statement\}\\
Please think step by step about whether the above statement is True or False, and put your final answer within $\backslash\backslash$boxed\{\}.
\end{tcolorbox}

\paragraph{Evaluation metrics of counter-example based backward reasoning} In the evaluation of counterexample-driven backward reasoning, we go beyond computing the F1 score for judging the truthfulness of a given statement. Inspired by the CounterMATH framework, we also assess GPT-4o’s capability to generate examples within its reasoning output. This is achieved through Example Extraction, which detects and retrieves instances where the model explicitly introduces or references counterexamples to support its claims. Alignment Assessment then determines whether each extracted example is consistent with a predefined Reference Example in terms of logical reasoning pattern, problem decomposition strategy, and goal relevance. Notably, since a proposition may have multiple valid counterexamples, exact replication of the reference is not mandatory for determining consistency. Instead, the reference serves as a guiding benchmark for GPT-4o, mitigating the risk of fully autonomous evaluations that may diverge from human-aligned reasoning standards. 

Specifically, the Examples metric indicates the percentage of problems where the model incorporates examples in its solution process. Strict Align represents the fraction of model-generated examples that fully match the reference in reasoning alignment, while Loose Align captures the proportion of cases where at least one example aligns with the reference example.

\section{Additional experiment and analysis of impact of atomic abilities across difficulty levels within the same field}

\subsection{Additional Results in Field Interaction}
We run each experiment multiple times with different seeds and report means and standard deviations as shown in Table \ref{tab:field_std}.

\begin{table}[h]
\centering
\caption{Performance comparison across various mathematical fields.}
\vspace{2.5mm}
\resizebox{\linewidth}{!}{
\begin{tabular}{l c c c c c c c c}
\toprule
Field & \multicolumn{2}{c}{Algebra} & \multicolumn{2}{c}{Analysis} & \multicolumn{2}{c}{Geometry} & \multicolumn{2}{c}{Topology} \\
\cmidrule(lr){2-3} \cmidrule(lr){4-5} \cmidrule(lr){6-7} \cmidrule(lr){8-9}
 Level & Low & High & Low & High & Low & High& Low & High \\
\midrule
Qwen-base & 80.5 & 65.2 & 67.7 & 66.5 & 52.1 & 53.4 & 52.1 & 53.4 \\
\midrule
Qwen-train-Algebra & 80.2 ($\pm $0.4) & 69.7 ($\pm $0.5) & 75.8 ($\pm $0.6) & 71.5 ($\pm $0.7) & 65.7 ($\pm $0.3) & 57.5 ($\pm $0.6) & 56.0 ($\pm $0.5) & 62.3 ($\pm $0.9) \\
Qwen-train-Geometry & 79.6 ($\pm $0.5) & 68.1 ($\pm $0.4) & 69.8 ($\pm $0.6) & 59.5 ($\pm $0.8) & 57.3 ($\pm $0.6) & 56.0 ($\pm $0.5) & 52.8 ($\pm $0.4) & 60.9 ($\pm $0.3) \\
\bottomrule
\end{tabular}}
\label{tab:field_std}
\end{table}

\subsection{Difficulty Interaction in Field Atomic Ability}\label{app:difficulty}

We conduct comparative experiments within each mathematical field to investigate how training on datasets of varying difficulty levels affects the development of atomic abilities. The results in Table \ref{tab:level_difficulty} demonstrate that models trained on \textbf{high-difficulty data} exhibit performance improvements on both easy and hard test, with particularly notable gains on the easier tasks. This indicates that mastering complex knowledge not only activates high-level atomic abilities but also facilitates the co-activation of lower-level atomic abilities. However, training solely on low-difficulty data can negatively affect performance on harder tasks. This phenomenon is especially evident in fields such as \textit{Algebra, Analysis, and Topology}, where models trained on easier data show a performance decline when evaluated on more difficult tasks. These findings suggest the need for careful \textbf{balancing of difficulty distribution} in training datasets to stimulate atomic abilities across the full spectrum and prevent \textbf{knowledge forgetting}.

\begin{table}[h]
\centering
\caption{Performance comparison of different training levels across various mathematical fields. We color the \colorbox{uparrowcolor}{positive\uparrowcolor{}} / \colorbox{downarrowcolor}{negative\downarrowcolor{}} influence as \colorbox{uparrowcolor}{green} / \colorbox{downarrowcolor}{red}.}
\vspace{2.5mm}
\resizebox{\linewidth}{!}{
\begin{tabular}{l c c c c c c c c}
\toprule
Field & \multicolumn{2}{c}{Algebra} & \multicolumn{2}{c}{Analysis} & \multicolumn{2}{c}{Geometry} & \multicolumn{2}{c}{Topology} \\
\cmidrule(lr){2-3} \cmidrule(lr){4-5} \cmidrule(lr){6-7} \cmidrule(lr){8-9}
 Level & Low & High & Low & High & Low & High& Low & High \\
\midrule
Qwen-base & 80.5 & 65.2 & 67.7 & 66.5 & 52.1 & 53.4 & 52.1 & 53.4 \\
\midrule
Qwen-train-l2 & 83.2 (\uparrowcolor{2.7}) & 68.3 (\uparrowcolor{3.1}) & 76.8 (\uparrowcolor{9.1}) & 66.4 (\downarrowcolor{0.1}) & 59.6 (\uparrowcolor{7.5}) & 57.7 (\uparrowcolor{4.3}) & 53.1 (\uparrowcolor{1.0}) & 53.8 (\uparrowcolor{0.4}) \\
Qwen-train-l1 & 79.5 (\downarrowcolor{1.0}) & 64.9 (\downarrowcolor{0.3}) & 70.5 (\uparrowcolor{2.8}) & 62.2 (\downarrowcolor{4.3}) & 58.9 (\uparrowcolor{6.8}) & 57 (\uparrowcolor{3.6}) & 52.2 (\uparrowcolor{0.1}) & 52.1 (\downarrowcolor{1.3}) \\
Qwen-train-all & 80 (\downarrowcolor{0.5}) & 69.7 (\uparrowcolor{4.5}) & 71.8 (\uparrowcolor{4.1}) & 64.9 (\downarrowcolor{1.6}) & 57.4 (\uparrowcolor{5.3}) & 56.1 (\uparrowcolor{2.7}) & 53.7 (\uparrowcolor{1.6}) & 59.1 (\uparrowcolor{5.7}) \\
\bottomrule
\end{tabular}}
\label{tab:level_difficulty}
\end{table}

\begin{table}[h]
\centering
\caption{Performance comparison of different training levels across various mathematical fields. We color the \colorbox{uparrowcolor}{positive\uparrowcolor{}} / \colorbox{downarrowcolor}{negative\downarrowcolor{}} influence as \colorbox{uparrowcolor}{green} / \colorbox{downarrowcolor}{red}.}
\vspace{2.5mm}
\resizebox{\linewidth}{!}{
\begin{tabular}{l c c c c c c c c}
\toprule
Field & \multicolumn{2}{c}{Algebra} & \multicolumn{2}{c}{Analysis} & \multicolumn{2}{c}{Geometry} & \multicolumn{2}{c}{Topology} \\
\cmidrule(lr){2-3} \cmidrule(lr){4-5} \cmidrule(lr){6-7} \cmidrule(lr){8-9}
 Level & Low & High & Low & High & Low & High& Low & High \\
\midrule
Qwen-base & 80.5 & 65.2 & 67.7 & 66.5 & 52.1 & 53.4 & 52.1 & 53.4 \\
\midrule
Qwen-train-l2 & 83.2 (\uparrowcolor{2.7}) & 68.3 (\uparrowcolor{3.1}) & 76.8 (\uparrowcolor{9.1}) & 66.4 (\downarrowcolor{0.1}) & 59.6 (\uparrowcolor{7.5}) & 57.7 (\uparrowcolor{4.3}) & 53.1 (\uparrowcolor{1.0}) & 53.8 (\uparrowcolor{0.4}) \\
Qwen-train-l1 & 79.5 (\downarrowcolor{1.0}) & 64.9 (\downarrowcolor{0.3}) & 70.5 (\uparrowcolor{2.8}) & 62.2 (\downarrowcolor{4.3}) & 58.9 (\uparrowcolor{6.8}) & 57 (\uparrowcolor{3.6}) & 52.2 (\uparrowcolor{0.1}) & 52.1 (\downarrowcolor{1.3}) \\
Qwen-train-all & 80 (\downarrowcolor{0.5}) & 69.7 (\uparrowcolor{4.5}) & 71.8 (\uparrowcolor{4.1}) & 64.9 (\downarrowcolor{1.6}) & 57.4 (\uparrowcolor{5.3}) & 56.1 (\uparrowcolor{2.7}) & 53.7 (\uparrowcolor{1.6}) & 59.1 (\uparrowcolor{5.7}) \\
\bottomrule
\end{tabular}}
\label{tab:level_difficulty}
\end{table}

\section{Human Evaluation}
We conduct a human evaluation on a sample of the data during the initial phase of our experiments in order to verify the accuracy of our automatic evaluation metrics. To do this, we selected two models, Gemini and Qwen, and sampled 20 problems from each dataset for manual review. Furthermore, since the majority of the problems are calculation-based, the final answers have a relatively fixed format. This, combined with our constraint that the answer must be in the \\boxed{} format, greatly simplified the validation process and made it easier to check for accuracy. The table below shows the number of cases (out of 20) where the automatic evaluation aligned with the human evaluation for each model and dataset.

\begin{table}[h]
\centering
\caption{The human evaluation category-wise scores.}
\label{tab:math-category-scores}
\resizebox{\linewidth}{!}{
\begin{tabular}{l cc cc cc cc c c c}
\toprule
\multirow{2}{*}{\textbf{Model}} & \multicolumn{2}{c}{\textbf{Algebra}} &
  \multicolumn{2}{c}{\textbf{Geometry}} &
  \multicolumn{2}{c}{\textbf{Analysis}} &
  \multicolumn{2}{c}{\textbf{Topology}} &
  \multirow{2}{*}{\textbf{Attr.}} & \multirow{2}{*}{\textbf{Def.}} & \multirow{2}{*}{\textbf{Forward Rea.}} \\
\cmidrule(lr){2-3}\cmidrule(lr){4-5}\cmidrule(lr){6-7}\cmidrule(lr){8-9}
 & l1 & l2 & l1 & l2 & l1 & l2 & l1 & l2 & & & \\
\midrule
Qwen2.5-Math-Instruct-7B
  & 20/20 & 20/20
  & 20/20 & 20/20
  & 20/20 & 20/20
  & 19/20 & 19/20
  & 18/20 & 20/20 & 20/20 \\
Gemini2.5-pro
  & 20/20 & 20/20
  & 20/20 & 20/20
  & 19/20 & 20/20
  & 20/20 & 20/20
  & 19/20 & 20/20 & 20/20 \\
\bottomrule
\end{tabular}}
\end{table}

\section{Visualization results of field atomic ability interaction}\label{app:heatmap}

\begin{figure}[h]
    \centering
    \includegraphics[width=0.5\linewidth]{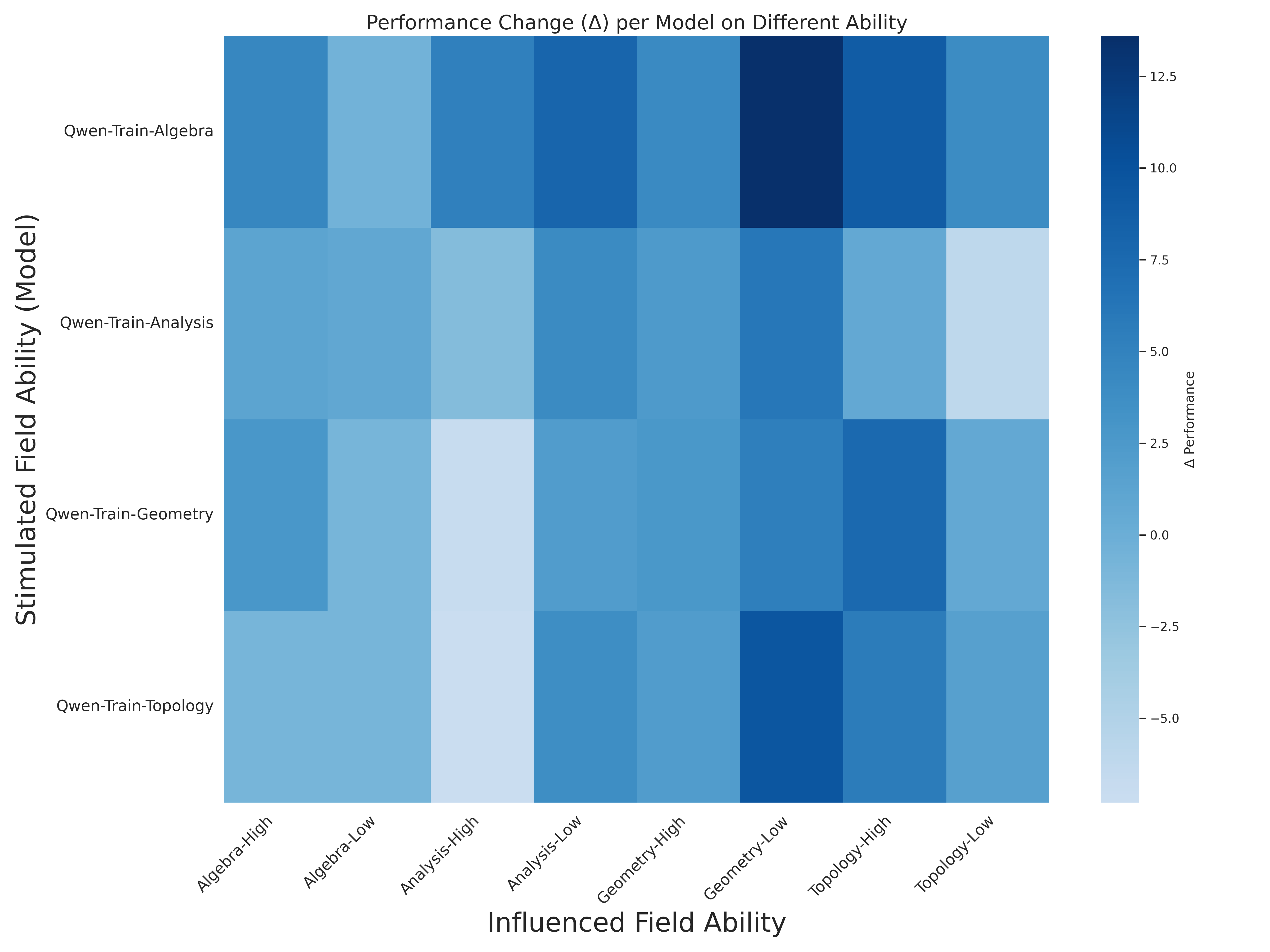}
    \caption{This figure shows the impact of stimulating one ability on the remaining abilities, where deeper colors indicate a greater positive facilitation effect, while lighter colors indicate a greater negative impact.}
    \label{fig:heat}
\end{figure}

We have provided a heat map in Figure \ref{fig:heat} that represents the correlation between atomic capabilities in the field, providing a more intuitive result.

\newpage

\end{document}